\documentclass[10pt,twocolumn,letterpaper]{article}

\usepackage[pagenumbers]{wacv} 

\definecolor{wacvblue}{rgb}{0.21,0.49,0.74}
\usepackage{multirow}
\usepackage[hidelinks]{hyperref}

\title{CSF-Net: Context-Semantic Fusion Network for Large Mask Inpainting}
\author{
Chae-Yeon Heo and Yeong-Jun Cho\thanks{Corresponding author}\\
Department of Artificial Intelligence Convergence\\
Chonnam National University, Gwangju, South Korea\\
{\tt\small \{cyheo001, yj.cho\}@jnu.ac.kr}
}

\begin{document}
\maketitle
\begin{abstract}
In this paper, we propose a semantic-guided framework to address the challenging problem of large-mask image inpainting, where essential visual content is missing and contextual cues are limited.
To compensate for the limited context, we leverage a pretrained Amodal Completion (AC) model to generate structure-aware candidates that serve as semantic priors for the missing regions.
We introduce Context-Semantic Fusion Network (CSF-Net), a transformer-based fusion framework that fuses these candidates with contextual features to produce a semantic guidance image for image inpainting.
This guidance improves inpainting quality by promoting structural accuracy and semantic consistency. CSF-Net can be seamlessly integrated into existing inpainting models without architectural changes and consistently enhances performance across diverse masking conditions. Extensive experiments on the \texttt{Places365} and \texttt{COCOA} datasets demonstrate that CSF-Net effectively reduces object hallucination while enhancing visual realism and semantic alignment. The code for CSF-Net is available at \href{https://github.com/chaeyeonheo/CSF-Net.git}{https://github.com/chaeyeonheo/CSF-Net.git}
\\
\end{abstract}    
\begin{figure}[t!]
\centering
\includegraphics[width=\columnwidth]{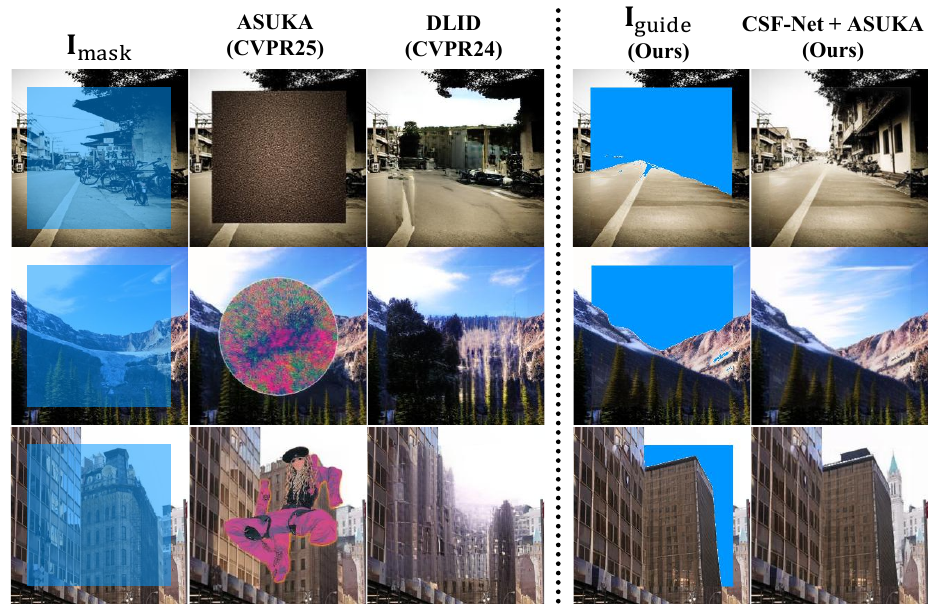}
\caption{
Large-mask inpainting comparison. 
Existing methods~\cite{Wang_2025_CVPR,Chen_2024_CVPR} often produce structural errors or object hallucinations in challenging cases. 
Incorporating our semantic guidance ($\mathbf{I}_{\text{guide}}$) yields more accurate and semantically coherent inpainting results.
}
\label{fig:large mask problem}
\end{figure}

\section{Introduction}
\label{sec:1.Introduction}

Image inpainting is an important vision task that restores missing regions with semantic consistency to the surrounding context.
It plays a crucial role in various applications, including photo editing, object removal, and scene reconstruction.
Early approaches relied on low-level cues such as color similarity and texture propagation, but struggled with large or complex missing regions due to limited semantic understanding.
Recent advances in generative models, such as Generative Adversarial Networks (GANs) and diffusion models, have significantly improved image inpainting.
Among these, GAN-based methods~\cite{Pathak_2016_CVPR,Nazeri_2019_ICCV} have enhanced inpainting by learning semantic structures and incorporating structural priors, leading to more coherent and context-aware results.
Diffusion-based approaches~\cite{Wang_2025_CVPR,Corneanu_2024_WACV} have further advanced inpainting by synthesizing detailed textures and improving inference efficiency.

However, despite recent progress, inpainting models still struggle in large-mask scenarios where much of the image is missing and contextual cues are limited.
As shown in Fig.~\ref{fig:large mask problem}, even state-of-the-art models often fail to generate semantically accurate content when key structural regions are missing.
This results in object hallucination, where the generated content appears visually plausible but is inconsistent with the surrounding context due to insufficient information about the missing areas.

To address this challenge, we incorporate semantic guidance by leveraging a pretrained amodal completion model~\cite{Ozguroglu_2024_CVPR}, which generates structure-aware candidates for the missing regions through object-level reasoning.
These candidates recover both shape and appearance, providing strong priors that guide semantically coherent inpainting.
Since the model learns to reason at the object level, it completes missing regions by extending only the visible parts of each object.
This prevents the generation of unrelated content and ensures that the results remain semantically consistent with each object instance.

\begin{figure}[t]
\centering
\includegraphics[width=\columnwidth]{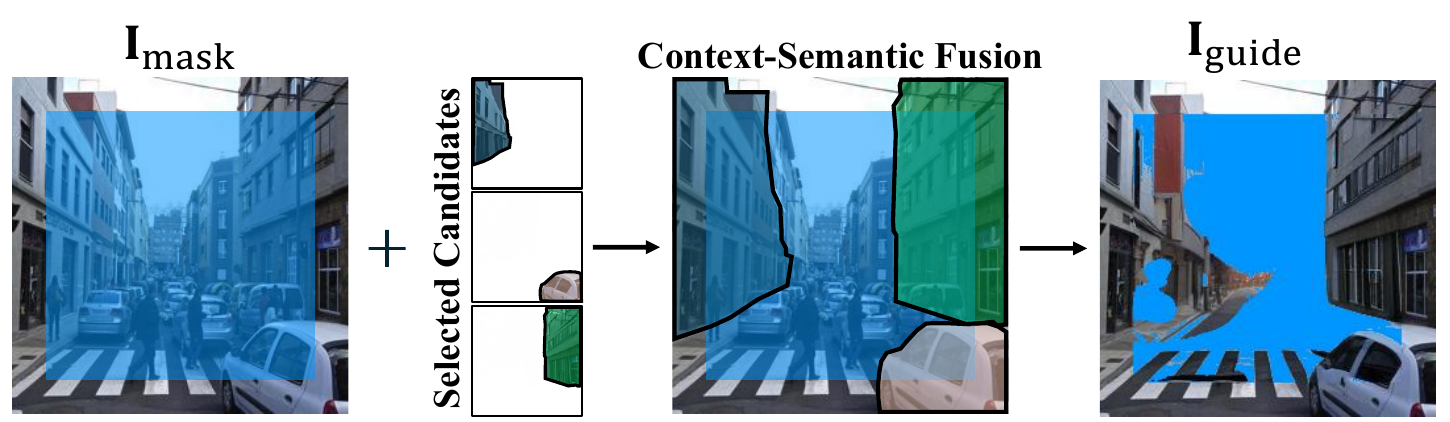}
\caption{
Overview of semantic guidance image ($\mathbf{I}_{\text{guide}}$) generation of CSF-Net.
This image incorporates object-level semantic priors and serves as an input to the inpainting model.
}
\label{fig:amodal_completion_concept}
\end{figure}

In this paper, we propose the Context-Semantic Fusion Network (CSF-Net), a transformer-based framework that generates a semantic guidance image to support structure-aware inpainting.
CSF-Net consists of three main components: a candidate generation module that selects plausible completions from amodal outputs (Sec.~\ref{sec:4.1.Methods_Candidate_Generation}); a dual encoder and a fusion decoder that extract and integrate contextual and semantic features (Sec.~\ref{sec:4.2.Methods_CSF-Transformer}); and a pixel selection module (Sec.~\ref{sec:4.3.Methods_Pixel_Selection}) that selects the final pixel values to construct the semantic guidance image.
As illustrated in Fig.\ref{fig:amodal_completion_concept}, the masked image $\mathbf{I}_{\text{mask}}$ and structure-aware candidates are encoded and fused to produce the semantic guidance image $\mathbf{I}_{\text{guide}}$.
This guidance provides high-level structural cues, enabling the inpainting model to restore missing regions with greater semantic accuracy and visual consistency.

We evaluate CSF-Net on the \texttt{Places365}~\cite{zhou2017places} and \texttt{COCOA}~\cite{zhu2017semantic} datasets under challenging masking scenarios, including Center Box (50\%, 80\%) and RandomBrush (50--80\%).
Extensive experiments confirm that CSF-Net consistently improves the performance of diverse inpainting models, including LaMa~\cite{Suvorov_2022_WACV}, MAT~\cite{Li_2022_CVPR}, 
DLID~\cite{Chen_2024_CVPR}, and ASUKA~\cite{Wang_2025_CVPR}.  
Our method achieves consistent gains across multiple metrics (FID, LPIPS, C@m), demonstrating both the generality and effectiveness of the proposed approach. 
Importantly, CSF-Net can be seamlessly integrated into existing inpainting architectures without requiring any modifications, making it a versatile enhancement module.

Our main contributions are as follows:
\begin{itemize}
    \item We introduce a semantic guidance strategy that leverages amodal completion to compensate for missing contextual cues in large-mask inpainting.         
    \item We propose CSF-Net, a transformer-based fusion framework that unifies contextual cues and semantic priors to generate the a semantic guidance image for inpainting.
    \item We show that CSF-Net improves the performance of diverse inpainting baselines under various masking conditions, without requiring any architectural modifications.
\end{itemize}
To the best of our knowledge, CSF-Net is the first attempt to leverage amodal completion for guiding image inpainting.

\section{Related Works}
\label{sec:2.Related_Works}
\paragraph{Image Inpainting.}
Image inpainting aims to restore missing or corrupted regions in images with content that is both visually coherent and semantically meaningful.
Traditional approaches such as~\cite{bertalmio2000image, criminisi2004region} propagated neighboring information or copied similar patches to fill in the gaps.
Structure propagation methods extended edges or contours from surrounding regions to fill missing areas.
Patch-based techniques copied similar patches from known regions into the target holes.
While effective for small or textured regions, these methods often produced blurry or repetitive artifacts in large, semantically complex areas.

Generative Adversarial Networks (GANs) \cite{NIPS2014_f033ed80} significantly advanced image inpainting by enabling the generation of semantically coherent content in missing regions.
Context Encoders~\cite{Pathak_2016_CVPR} first applied GANs to predict missing regions from high-level semantic features.
Later methods such as EdgeConnect~\cite{Nazeri_2019_ICCV} and StructureFlow~\cite{ren2019structureflow} incorporated structural priors, including edges and semantic layouts, to guide the inpainting process.
These works demonstrated the benefits of guiding inpainting with explicit structure.

More recently, MaskGIT~\cite{Chang_2022_CVPR} improved inference speed through parallel iterative decoding, offering an efficient alternative to autoregressive models.
Diffusion model~\cite{sohl2015deep,rombach2022high} has recently shown strong performance in image synthesis but struggle with artifacts near mask boundaries in large-mask settings.
To address this, a post-processing method, ASUKA~\cite{Wang_2025_CVPR} applies pretrained priors and decoding to reduce hallucinations and color inconsistencies.
LatentPaint~\cite{Corneanu_2024_WACV} further improves upon this by performing inpainting directly in the latent space of pretrained diffusion models.
Nonetheless, generative approaches remain limited when the visible context lacks sufficient semantic cues.

Parallel to the development of generative models such as GANs and diffusion, several methods have been proposed to address challenges of large-mask image inpainting.
LaMa~\cite{Suvorov_2022_WACV} employs Fast Fourier Convolutions (FFC) to establish global receptive fields early in the network, facilitating the completion of repetitive patterns.
Transformer-based methods such as MAT~\cite{Li_2022_CVPR} and DLID~\cite{Chen_2024_CVPR} further improve performance by modeling long-range dependencies and learning latent codes from visible regions.
However, these models still rely heavily on visible context, resulting in limited performance when semantic cues are sparse~\cite{zheng2019pluralistic}.

\begin{figure*}[t!]
\centering
\includegraphics[width=1\textwidth]{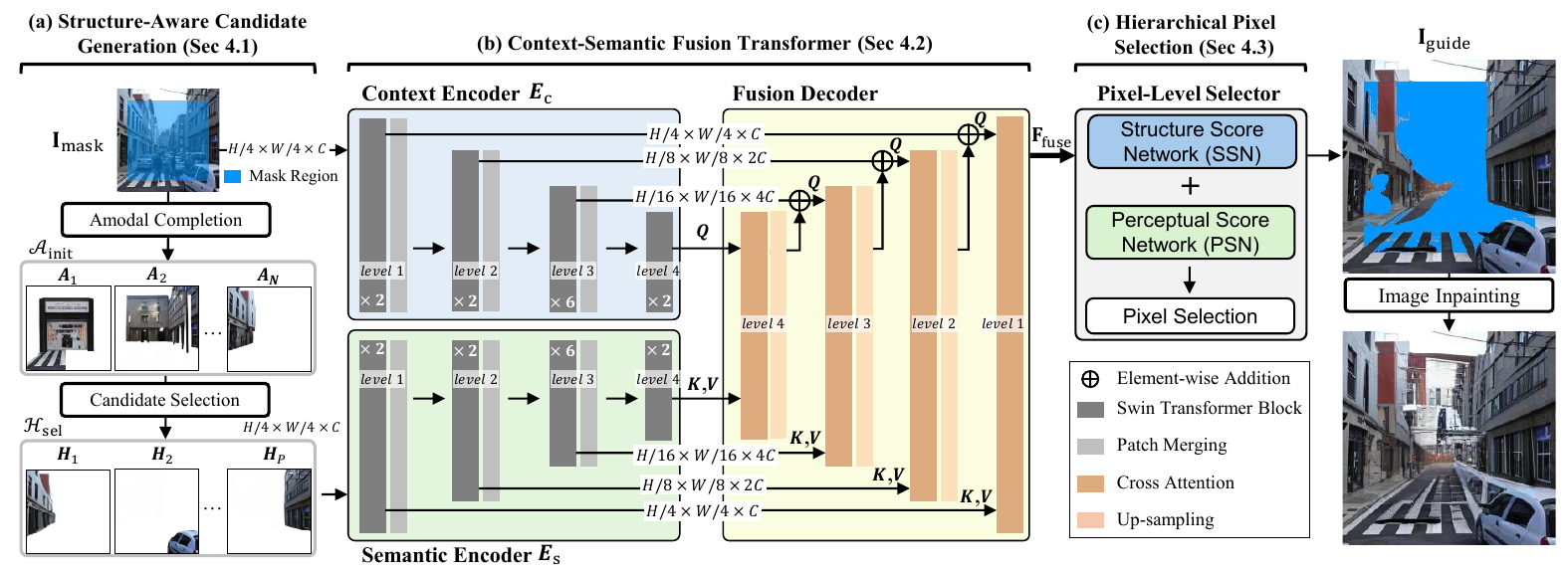}
  \caption{
  Overview of the proposed CSF-Net. 
  (a) A pretrained amodal completion model generates multiple object completions, and context-inconsistent candidates are filtered out. 
  (b) Dual Swin-Transformer encoders extract multi-scale features from the masked image and selected candidates, which are fused via a cross-attention fusion decoder. 
  (c) Hierarchical pixel selection is performed using structural and perceptual scores to generate the final semantic guidance image $\mathbf{I}_{\text{guide}}$.
  }
\label{fig:csf_architecture}
\end{figure*}

\paragraph{Amodal Completion.}
Amodal completion aims to infer the full shape and appearance of partially occluded objects, including their invisible parts. 
In contrast to inpainting methods that fill arbitrary masks, amodal completion focuses on object-level reasoning to reconstruct complete entities from partial observations.
Early methods usually relied on geometric cues, while deep learning approaches predicted complete binary masks~\cite{li2016amodal}, focusing primarily on shape recovery.
Built on generative modeling, SeGAN~\cite{Ehsani_2018_CVPR} jointly performed segmentation and appearance synthesis.
More recently, Pix2Gestalt~\cite{Ozguroglu_2024_CVPR} employs diffusion models to synthesize complete object appearances using large-scale pretrained generative priors.

In this paper, we exploit this capability to generate complete object-level representations, providing high-level semantic guidance for inpainting.
This enables our method to overcome the limitations of context-only models, particularly in large-mask scenarios.

\section{Motivation and Main Ideas}
\label{sec:3.Motivation}
The large mask image inpainting problem is inherently ill-posed, especially when large regions are missing and lack reliable visual cues.
In such cases, recent diffusion-based inpainting methods~\cite{Lugmayr_2022_CVPR,Wang_2025_CVPR} often hallucinate content that is semantically inconsistent with the surrounding context.
This is mainly due to their training objective, which prioritizes generating visually plausible textures over understanding high-level semantics.
As a result, their performance significantly degrades in large missing areas, where accurate content prediction requires deeper semantic reasoning.

To address this limitation, we propose the Context-Semantic Fusion Network (CSF-Net), which guides the inpainting process using semantic guidances.
Conventional inpainting models often fill missing regions by relying only on the surrounding visible pixels, without reasoning about the underlying object structure.
In contrast, our approach leverages a pretrained Amodal Completion (AC) model~\cite{Ozguroglu_2024_CVPR} that learns to infer the complete shape of partially occluded objects from visible cues.
This model generates multiple completions that reflect strong object-level semantics. 
Among these, only contextually consistent candidates are selected and then fused to form a coherent semantic representation.
This fused result, denoted as $\mathbf{I}_{\text{guide}}$, serves as a guidance for image inpainting.

\section{Proposed Methods}
\label{sec:4.Methods}
The proposed CSF-Net generates a guidance image $\mathbf{I}_{\text{guide}}$, which can serve as structural guidance for existing image inpainting methods.
The overall framework consists of three stages as shown in Fig.~\ref{fig:csf_architecture}.
First, we generate structure-aware semantic candidates by leveraging a pretrained Amodal Completion (AC) model that infers object shapes from visible regions (Sec. 4.1).
Next, we fuse features from the masked image and the semantic candidates through a pair of encoders and a fusion decoder (Sec. 4.2).
Finally, we select the optimal candidate for each pixel to compose $\mathbf{I}_{\text{guide}}$ using a multi-scale pixel selection (Sec. 4.3).

\subsection{Structure-Aware Candidate Generation}
\label{sec:4.1.Methods_Candidate_Generation}
In this section, we generate multiple amodal completions using a pretrained AC model.
We then evaluate each completion based on its structural and contextual consistency with the visible region, and select only the reliable ones as semantic candidates as shown in Fig.~\ref{fig:csf_architecture}(a). 
The AC model was originally developed to reconstruct the invisible parts of partially occluded objects in natural scenes.
We apply it to the inpainting task by treating the artificial mask as an occluder and reconstructing the occluded structures near its boundary.
Given a masked image $\mathbf{I}_{\text{mask}}$, the AC model generates an initial set of $N$ completions, denoted by $\mathcal{A}_{\text{init}} = \{ \mathbf{A}_i \}_{i=1}^{N}$, while preserving candidate diversity through its segment-wise generation. At this stage, the visible areas are pre-segmented and each segment is completed individually, ensuring that structurally distinct completion candidates can emerge.
Each completion $\mathbf{A}_i$ serves as a structurally plausible hypothesis for the missing region. However, some completions are semantically inconsistent with the visible context, which can degrade the inpainting quality when included in the fusion.

To mitigate this, we evaluate initial completions using a combined consistency score, 
defined as $S_{\text{valid}} = S_{\text{MSE}} + S_{\text{LPIPS}}$, 
where $S_{\text{MSE}}$ measures pixel-wise similarity and $S_{\text{LPIPS}}$ quantifies perceptual similarity based on the LPIPS metric~\cite{zhang2018unreasonable}%
\footnote{In practice, $S_{\text{MSE}}$ and $S_{\text{LPIPS}}$ are first computed as pixel-wise error and perceptual similarity (LPIPS), respectively, and then inverted. The two inverted values are summed to obtain the final score. Therefore, a higher $S_{\text{valid}}$ indicates greater consistency.}.
Each $\mathbf{A}_i$ is evaluated by comparing its content against the visible region of the masked image. 
The consistency score reflects how well each completion aligns with the contextual information contained in the visible region of $\mathbf{I}_{\text{mask}}$. 
Completions with pixel-wise or perceptual inconsistencies with the visible region receive lower scores.
We rank all completions by their $S_{\text{valid}}$ scores and select the top $P$ most consistent ones to form a selected set $\mathcal{H}
_{\text{sel}} =  \{ \mathbf{H}_i \}_{i=1}^{P}$, referred to semantic candidates for inpainting.
This selection process reduces semantic ambiguity and narrows the candidate space for more useful fusion.

\subsection{Context-Semantic Fusion Transformer}
\label{sec:4.2.Methods_CSF-Transformer}
The proposed CSF-Net integrates two sources of information: a masked image $\mathbf{I}_{\text{mask}}$ and a set of selected semantic candidates $\mathcal{H}_{\text{sel}}$.
It aims to produce a fused representation that jointly captures structural and semantic cues. 
To process these inputs, CSF-Net employs a dual-encoder architecture based on Swin Transformer blocks~\cite{liu2021swin}, followed by a fusion decoder.
The overall structure of CSF-Net is designed to follow a U-Net–style~\cite{ronneberger2015u} architecture with multi-resolution skip connections, enabling the network to capture long-range dependencies while preserving fine-grained spatial details.
An overview architecture is shown in Fig.~\ref{fig:csf_architecture}(b).

\paragraph{Context and Semantic Encoders.}
The encoders in CSF-Net independently process the masked input image $\mathbf{I}_{\text{mask}}$ and the semantic candidates $\mathcal{H}_{\text{sel}}$, extracting multi-scale features.
The masked image is passed through the context encoder $E_c$ to extract contextual features at different resolution levels:
\begin{equation}
\mathbf{F}_{\text{ctx}}^{(l)} = E_c\left(\mathbf{I}_{\text{mask}}\right)^{(l)},
\end{equation}
where $l=\{1,2,\dots,L\}$ denotes the resolution level within the encoder (e.g., from fine to coarse).
These features encode the global layout and boundary-sensitive structures of the visible region, providing spatial cues. 
In parallel, each semantic candidate $\mathbf{H}_i$ is passed through the semantic encoder $E_s$ as follows:
\begin{equation}
\mathbf{F}_{\text{sem},i}^{(l)} = E_s\left(\mathbf{H}_i\right)^{(l)}.
\end{equation}
Both $E_c$ and $E_s$ adopt the same Swin Transformer architecture, composed of stacked transformer blocks and patch merging. These layers gradually downsample the features, enabling the model to capture multi-scale representations essential for reconstructing missing regions. 

\paragraph{Fusion Decoder.}
The fusion decoder performs hierarchical fusion from coarse to fine resolutions, starting at level the final level $L$.
At this level, the context feature serves as the query, while the semantic features from all candidates serve as the keys and values in a cross-attention:
\begin{equation}
\mathbf{F}^{(L)}_{\text{fuse}} = \text{CrossAttn}\left(Q = \mathbf{F}^{(L)}_{\text{ctx}},\ K = V = \mathbf{F}^{(L)}_{\text{sem},i} \right).
\end{equation}
Here, $\mathbf{F}^{(L)}_{\text{sem},i}$ denotes the semantic features from each candidate, treated as individual tokens in the attention.
This initial output of the decoder is a set of fused feature maps that unify contextual and semantic information across multiple scales.

For finer resolution levels $l = \{L-1, L-2, \ldots, 1\}$, the decoder upsamples the fused output from the previous level and adds it element-wise to the corresponding context feature to form the query:

{\footnotesize
\begin{equation}
\scalebox{0.84}{$
\mathbf{F}^{(l)}_{\text{fuse}} = \text{CrossAttn}\left(Q = U(\mathbf{F}_\text{fuse}^{(l+1)}) + \mathbf{F}_{\text{ctx}}^{(l)},\ K = V = \mathbf{F}^{(l)}_{\text{sem},i} \right),
$}
\end{equation}
}
where $U(\cdot)$ denotes bilinear upsampling.
These fused features $\mathbf{F}^{(l)}_{\text{fuse}}$ serve as a strong representation for selecting plausible pixels in the masked regions during the final reconstruction stage.

\subsection{Hierarchical Pixel Selection}
\label{sec:4.3.Methods_Pixel_Selection}
To generate the semantic guidance image $\mathbf{I}_{\text{guide}}$, we introduce a hierarchical pixel selection that determines the most plausible content for each masked pixel. This module selects pixel values from the candidate set ${\mathbf{H}_i}$ based on their consistency with the fused feature representation.
Each semantic candidate $\mathbf{H}_i$ has the same spatial resolution as the input image but contains valid pixel values only within its completed region, while the rest remains undefined.
Because multiple candidates may overlap at the same pixel location, the model must evaluate which candidate provides the most suitable content for each pixel.
To address this, the semantic candidates $\mathbf{H}_i$ are evaluated in a coarse-to-fine manner across multiple resolution levels $l$.

\begin{figure}[t!]
\centering
\includegraphics[width=\columnwidth]{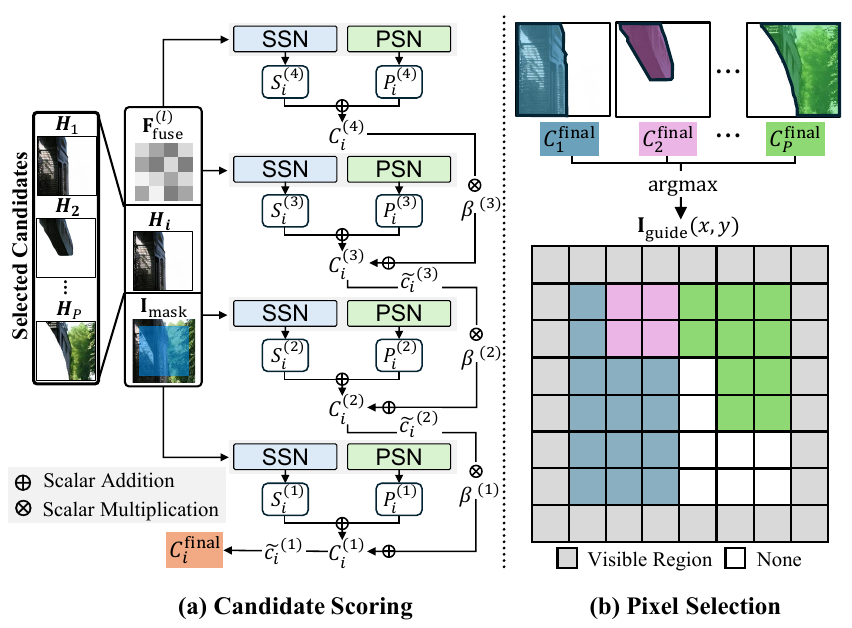}
\caption{
Hierarchical Pixel Selection in the CSF-Net. 
(a) The Structure Score Network (SSN) and Perceptual Score Network (PSN) compute confidence scores at each scale using fused features and the masked input. Multi-scale consistency is enforced via learnable coefficients $\beta$. 
(b) At the finest scale, the highest-scoring candidate is selected for each pixel to form the semantic guidance image $\mathbf{I}_{\text{guide}}$. 
}
\label{fig:pixel_selection_network}
\end{figure}

The masked image $\mathbf{I}_{\text{mask}}$ serves as a contextual reference, providing spatial and structural cues for evaluating semantic candidates.
To assess the quality of each candidate, we introduce two complementary scoring networks, as shown in Fig.~\ref{fig:pixel_selection_network}(a).
The Structure Score Network (SSN), implemented as a lightweight convolutional network, estimates structural plausibility. 
In parallel, the Perceptual Score Network (PSN) leverages a pretrained VGG-19 ~\cite{johnson2016perceptual} to evaluate perceptual quality and texture realism.
The score maps for each network are computed as follows:

{\footnotesize
\begin{equation}
S_{i}^{(l)}(x, y) = \text{SSN}\left(\left[\mathbf{F}_{\text{fuse}}^{(l)}(x, y),  \mathbf{H}_i^{(l)}(x, y),  \mathbf{I}_{\text{mask}}^{(l)}(x, y)\right]\right),
\end{equation}
}
{\footnotesize
\begin{equation}
P_{i}^{(l)}(x, y) = \text{PSN}\left(\left[\mathbf{F}_{\text{fuse}}^{(l)}(x, y), \mathbf{H}_i^{(l)}(x, y),  \mathbf{I}_{\text{mask}}^{(l)}(x, y)\right]\right),
\end{equation}
}

where $(x, y)$ denotes pixel coordinates, and $i$ is the index of the candidate.
Both the maksed image $\mathbf{I}_{\text{mask}}$ and candidate completions $\mathbf{H}_i$ are downsampled to match the spatial resolution of the $l$-th level.

The aggregated score map for each candidate at each level is computed as the average of the structural and perceptual score maps:
\begin{equation}
C_i^{(l)}(x, y) = \frac{1}{2} \left(S_{i}^{(l)}(x, y) + P_{i}^{(l)}(x, y)\right) .
\end{equation}
To improve spatial consistency across resolution levels, we refine $C_i^{(l)}$ by blending it with the upsampled score from the coarser level $l+1$, weighted by a learnable $\beta^{(l)}$, as follows:
\begin{equation}
\scalebox{0.9}{$
\Tilde{C}_i^{(l)}(x, y) = (1 - \beta^{(l)}) C_i^{(l)}(x, y) + \beta^{(l)} U\left(C_i^{(l+1)}(x, y)\right),
$}
\end{equation}
where $U(\cdot)$ denotes bilinear upsampling. The coefficients $\beta^{(l)}$ are jointly optimized during training to adaptively control multi-scale information flow.
This refinement is applied recursively from coarse to fine levels, resulting in the final confidence score map $C_i^{\text{final}}(x, y)$ at the highest resolution.

To construct the semantic guidance image $\mathbf{I}_{\text{guide}}$, we perform discrete pixel-wise selection using the refined confidence scores $C_i^{\text{final}}(x, y)$.
At each location $(x, y)$, the model selects the candidate index with the highest confidence:
\begin{equation}
i^*(x, y) = \underset{i \in \{1,2,\ldots,P\}}{\mathrm{argmax}} C_i^{\text{final}}(x, y).
\end{equation}
If the highest score is below a predefined threshold, the pixel remains masked.
Using the selected indices $i^*(x, y)$, the final guidance image is assembled as:
\begin{equation}
\mathbf{I}_{\text{guide}}(x, y) = \mathbf{H}_{i^*(x,y)}(x, y).
\label{eq:iguide_selection}
\end{equation}
This selection strategy enables the model to choose the most suitable candidate for each pixel, even though the argmax operation in Eq.~\ref{eq:iguide_selection} is non-differentiable. The loss is computed on the final guidance image, and the gradients are propagated to the SSN and the fusion transformer, which learn to assign higher scores to consistent candidates.

\begin{table*}[t]
\centering
\setlength{\tabcolsep}{0.9mm}
\begin{tabular}{r|ccc|ccc|ccc|l}
\hline
\multirow{3}{*}{\textbf{Methods}} 
    & \multicolumn{9}{c|}{\textbf{\texttt{Places365}~\cite{zhou2017places}}} 
    & \multirow{3}{*}{\textbf{References}} \\ \cline{2-10}
    & \multicolumn{3}{c|}{\textbf{Center Box (80\%)}} 
    & \multicolumn{3}{c|}{\textbf{Center Box (50\%)}} 
    & \multicolumn{3}{c|}{\textbf{Random (50--80\%)}} & \\ 
\cline{2-10}
    & FID$\downarrow$ & LPIPS$\downarrow$ & C@m$\uparrow$ 
    & FID$\downarrow$ & LPIPS$\downarrow$ & C@m$\uparrow$ 
    & FID$\downarrow$ & LPIPS$\downarrow$ & C@m$\uparrow$ & \\
\hline
MaskGIT~\cite{Chang_2022_CVPR}     & 21.66      & 0.425             & 0.630             & 6.842             & 0.175             & 0.726              & 44.32             & 0.428             & 0.684             & CVPR 2022 \\
LaMa~\cite{Suvorov_2022_WACV}      & 16.72      & 0.346             & 0.688             & 5.858             & 0.131             & 0.777              & 11.55             & 0.355             & 0.743             & WACV 2022 \\
MAT~\cite{Li_2022_CVPR}            & 17.50      & 0.364             & 0.670             & 4.492             & 0.139             & 0.758              & 10.77             & 0.350             & 0.742             & CVPR 2022 \\
DLID~\cite{Chen_2024_CVPR}         & 26.16      & 0.405             & 0.680             & 6.660             & 0.148             & 0.795              & 17.12             & 0.384             & 0.725             & CVPR 2024 \\
ASUKA~\cite{Wang_2025_CVPR}        & 10.10      & 0.377             & 0.701             & 4.408             & 0.143             & 0.755              & 5.835             & 0.332             & 0.802             & CVPR 2025 \\
\hline
CSF-Net + ASUKA (Ours)  & \textbf{9.434}    & \textbf{0.332}    & \textbf{0.702}    & \textbf{3.612}    & \textbf{0.105}    & \textbf{0.796}     & \textbf{5.324}             & \textbf{0.325}             & \textbf{0.803}             & -- \\ 
\hline
\end{tabular}
\caption{Quantitative comparison between state-of-the-art inpainting methods and our CSF-enhanced model (shown with ASUKA integration) at $256 \times 256$ resolution on \texttt{Places365}. \textbf{Bold} indicates the best performance for each metric.}
\label{tab:base_vs_ours}
\end{table*}

\subsection{Loss Functions}
\label{sec:Methods_Loss_Functions}
Our model is trained end-to-end, jointly optimizing the Context-Semantic Feature Fusion Transformer (Sec.~\ref{sec:4.2.Methods_CSF-Transformer}) and the Hierarchical Pixel Selection (Sec.~\ref{sec:4.3.Methods_Pixel_Selection}). 
While the fused semantic feature map $\mathbf{F}_{\text{fuse}}$ is not directly supervised, it is implicitly guided by the loss applied to the guidance image $\mathbf{I}_{\text{guide}}$.

To ensure that the guidance image $\mathbf{I}_{\text{guide}}$ appears natural and aligns well with the ground truth in the filled regions, we define a reconstruction loss $\mathcal{L}_{\text{recon}}$.
Note that the loss is applied only to the pixels actually filled by the proposed CSF-Net within the masked area. It consists of three components as follows:
\begin{equation}
\mathcal{L}_{\text{recon}} = \mathcal{L}_1 + \mathcal{L}_{\text{perc}} + \mathcal{L}_{\text{smooth}},
\end{equation}
where $\mathcal{L}_1$ measures pixel-wise differences in the inpainted region,
$\mathcal{L}_{\text{perc}}$ encourages perceptual similarity to the ground truth by comparing deep features from a pretrained VGG network~\cite{johnson2016perceptual}.
$\mathcal{L}_{\text{smooth}}$ discourages abrupt or inconsistent candidate switching between neighboring pixels.
Overall, $\mathcal{L}_{\text{recon}}$ guides $\mathbf{I}_{\text{guide}}$ to closely match the ground truth in terms of both pixel accuracy and perceptual quality, while ensuring a smooth and coherent inpainting result.

To promote consistency across decoder levels, we introduce a hierarchical consistency loss $\mathcal{L}_{\text{hier}}$ that encourages finer-scale reconstructions to align with coarser-scale outputs:
\begin{equation}
\mathcal{L}_{\text{hier}} = \frac{1}{L-1} \sum_{l=1}^{L-1} \left\| \mathbf{I}^{(l)} - D( \mathbf{I}^{(l-1)}) \right\|_1,
\end{equation}
where $\mathbf{I}^{(l)}$ denotes the reconstructed guidance image at decoder level $l$, and $D\left( \mathbf{I}^{(l-1)} \right)$ represents the bilinearly downsampled output from the previous coarser level.
This loss encourages pixel-wise consistency across scales, helping the decoder produce stable and coherent results throughout the coarse-to-fine refinement process.

The total loss is defined as:
\begin{equation}
\mathcal{L}_{\text{total}} = \lambda \cdot \mathcal{L}_{\text{recon}} + (1 - \lambda) \cdot \mathcal{L}_{\text{hier}},
\end{equation}
where each term contributes to reconstruction quality and hierarchical consistency, respectively.

\section{Experimental Results}
\label{sec:experiment}

\subsection{Evaluation Settings}
\label{sec:Evaluation_Settings}
\paragraph{Datasets and Settings.}
We conduct our experiments on the \texttt{Places365}~\cite{zhou2017places}, a widely used scene-centric benchmark for image inpainting, and additionally evaluate on \texttt{COCOA}~\cite{zhu2017semantic} as a test set to verify the effectiveness of our method in object-centric scenarios.
Another representative benchmark is \texttt{CelebA-HQ} ~\cite{karras2018progressive}, which primarily consists of aligned human face images. 
Since our framework emphasizes object-level amodal completion, \texttt{CelebA-HQ} ~\cite{karras2018progressive} is not suitable for our setting. 
Therefore, we focus on \texttt{Places365} ~\cite{zhou2017places}, \texttt{COCOA}~\cite{zhu2017semantic}, which contain diverse scene categories with complex structures, providing a more appropriate testbed for evaluating our method. We randomly sample 30 images per category (365 categories in total), resulting in 10,950 training images. 
The official validation set of \texttt{Places365}~\cite{zhou2017places} and 1,450 images from \texttt{COCOA}~\cite{zhu2017semantic} are used as the evaluation and test sets, respectively.
All images are resized to a resolution of $256 \times 256$.
Since CSF-Net is designed to fuse outputs from pretrained models rather than training an entire network, it requires significantly fewer training samples compared to conventional end-to-end inpainting approaches.
We further define three mask settings for our experiments: (1) Center Box 50\%, (2) Center Box 80\%, and (3) RandomBrush masks ranging from 50–80\% area coverage with mixed brush strokes and rectangles, following the strategy used in MAT~\cite{Li_2022_CVPR}.
The Center Box masks simulate structured occlusions by masking a fixed-size box at the image center, while RandomBrush generates more irregular and organic missing regions.
We train our model using the AdamW optimizer with a learning rate of $2 \times 10^{-4}$, a batch size of 12, and 200 training epochs. All experiments are conducted on a single NVIDIA A100 GPU.

\begin{table}[t]
\small
\centering
\renewcommand{\arraystretch}{1.05}
\setlength{\tabcolsep}{0.9mm}

\begin{tabular}{r|ccc|ccc} 
\hline
\multirow{3}{*}{\textbf{Methods}} & \multicolumn{6}{c}{\textbf{\texttt{Places365} ~\cite{zhou2017places}}} \\ \cline{2-7} 
                 & \multicolumn{3}{c|}{\textbf{Random (50--80\%)}} 
                 & \multicolumn{3}{c}{\textbf{Center Box (80\%)}} \\
\cline{2-7}
& FID$\downarrow$ & LPIPS$\downarrow$ & C@m$\uparrow$ & FID$\downarrow$ & LPIPS$\downarrow$ & C@m$\uparrow$ \\
\hline
LaMa~\cite{Suvorov_2022_WACV}   & 11.55          & 0.355          & 0.743          & 16.72          & 0.346          & 0.688 \\
+CSF-Net(Ours) & \textbf{11.17} & \textbf{0.309} & \textbf{0.761} & \textbf{15.93} & \textbf{0.326} & \textbf{0.701} \\
\hline
MAT~\cite{Li_2022_CVPR}         & 10.77          & 0.350          & 0.742          & 17.50          & 0.364          & 0.670 \\
+CSF-Net(Ours) & \textbf{10.61} & \textbf{0.325} & \textbf{0.744} & \textbf{16.08} & \textbf{0.338} & \textbf{0.743} \\
\hline
DLID~\cite{Chen_2024_CVPR}      & 17.12          & 0.384          & 0.725          & 26.16          & 0.405          & 0.680 \\
+CSF-Net(Ours) & \textbf{15.45} & \textbf{0.362} & \textbf{0.732} & \textbf{21.13} & \textbf{0.400} & \textbf{0.693} \\
\hline
ASUKA~\cite{Wang_2025_CVPR}     & 5.835          & 0.332          & 0.802          & 10.10          & 0.377          & 0.701 \\
+CSF-Net(Ours) & \textbf{5.324} & \textbf{0.325} & \textbf{0.803} & \textbf{9.434} & \textbf{0.332} & \textbf{0.702} \\
\hline
\end{tabular}
\begin{tabular}{r|ccc|ccc} 
\hline
\multirow{3}{*}{\textbf{Methods}} & \multicolumn{6}{c}{\textbf{\texttt{COCOA}~\cite{zhu2017semantic}}} \\ \cline{2-7}
                 & \multicolumn{3}{c|}{\textbf{Random (50--80\%)}} 
                 & \multicolumn{3}{c}{\textbf{Center Box (80\%)}} \\
\cline{2-7}
& FID$\downarrow$ & LPIPS$\downarrow$ & C@m$\uparrow$ & FID$\downarrow$ & LPIPS$\downarrow$ & C@m$\uparrow$ \\
\hline
LaMa~\cite{Suvorov_2022_WACV}   & 64.98          & 0.375          & 0.719          & 83.56          & 0.428          & 0.653 \\
+CSF-Net(Ours) & \textbf{62.55} & \textbf{0.334} & \textbf{0.720} & \textbf{71.20} & \textbf{0.373} & \textbf{0.661} \\
\hline
MAT~\cite{Li_2022_CVPR}         & 60.24          & 0.351          & 0.723          & 78.37          & 0.459          & 0.652 \\
+CSF-Net(Ours) & \textbf{58.07} & \textbf{0.298} & \textbf{0.780} & \textbf{75.65} & \textbf{0.394} & \textbf{0.688} \\
\hline
DLID~\cite{Chen_2024_CVPR}      & 71.90          & 0.395          & 0.708          & 90.37          & 0.441          & 0.641 \\
+CSF-Net(Ours) & \textbf{63.93} & \textbf{0.373} & \textbf{0.742} & \textbf{72.04} & \textbf{0.417} & \textbf{0.665} \\
\hline
ASUKA~\cite{Wang_2025_CVPR}     & 69.77          & 0.446          & 0.742          & 50.90          & 0.399          & 0.666 \\
+CSF-Net(Ours) & \textbf{54.56} & \textbf{0.387} & \textbf{0.786} & \textbf{49.77} & \textbf{0.376} & \textbf{0.708} \\
\hline
\end{tabular}
\caption{
Performance comparison of models with CSF-Net integration on \texttt{Places365}~\cite{zhou2017places} and \texttt{COCOA}~\cite{zhu2017semantic}.
}
\label{tab:csf_two_datasets}
\end{table}

\begin{figure*}[t]
\centering
\includegraphics[width=\textwidth]{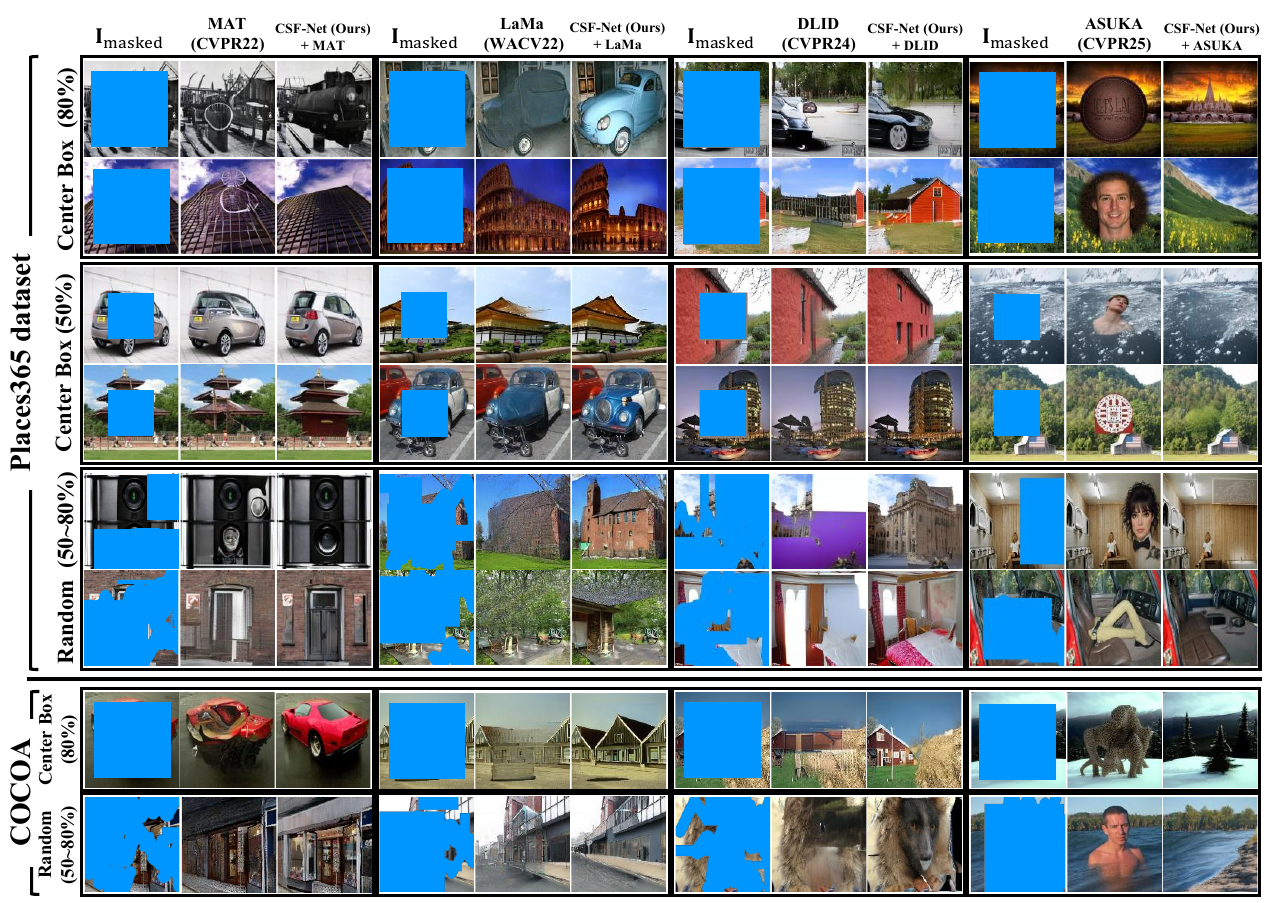}
\caption{ Comprehensive qualitative comparison under different mask configurations using the \texttt{Places365}~\cite{zhou2017places} (evaluated on all three mask types) and \texttt{COCOA}~\cite{zhu2017semantic} (evaluated on Center Box 80\% and RandomBrush 50--80\%).Our CSF-Net consistently generates clearer and more coherent results compared to baseline methods across diverse scenes and mask types, effectively reducing object hallucination.}
\label{fig:qualitative_256}
\end{figure*}

\paragraph{Evaluation Metrics.}
Our main evaluation objectives are to evaluate the visual quality, perceptual fidelity, and semantic consistency of the inpainted images.
To this end, we adopt the following three metrics: FID~\cite{heusel2017gans} to assess global realism, LPIPS~\cite{zhang2018unreasonable} for perceptual similarity, and C@m~\cite{Wang_2025_CVPR} for evaluating object-level semantic consistency within masked regions. 
C@m is a CLIP-based metric that measures the similarity between the restored region and the ground truth, effectively capturing object-level semantic consistency. 
As CSF-Net aims to restore semantically meaningful content within masked regions, C@m offers a more appropriate evaluation by measuring object-level alignment using CLIP-based features.
To complement the quantitative evaluation and provide a more comprehensive assessment, we present qualitative results illustrating visual fidelity and semantic plausibility of the inpainted outputs.

\subsection{Comparison with State-of-the-Art Methods}
\label{sec:Comparison_SOTA}
We compare CSF-Net with state-of-the-art inpainting methods.
Table~\ref{tab:base_vs_ours} presents the performance comparison on the \texttt{Places365} ~\cite{zhou2017places}.
While CSF-Net can be integrated into various inpainting frameworks, we adopt ASUKA as the baseline in this experiment.

To evaluate the general applicability of CSF-Net, we integrate it into four state-of-the-art inpainting models as shown in Tab~\ref{tab:csf_two_datasets}.  
The results demonstrate that performance gains are particularly notable under challenging Center Box masks (50\% and 80\%), while improvements with RandomBrush masks (50--80\%) are somewhat smaller than those for Center Box masks, due to their irregular patterns allowing richer use of surrounding context.  
Nevertheless, CSF-Net consistently improves the performance of all base models across various masking conditions and evaluation metrics.  
Note that the \texttt{COCOA}~\cite{zhu2017semantic} dataset contains relatively fewer samples, which leads to higher FID values overall. However, even under this limitation, CSF-Net yields consistent performance improvements, highlighting its robustness. 
These results demonstrate that CSF-Net provides effective object-aware guidance and consistently improves performance regardless of the underlying inpainting model.  
Moreover, a key advantage of CSF-Net is that it does not require any architectural modifications to the baseline inpainting frameworks.
The proposed method can be applied simply by generating a guidance image from the input mask.

Figure~\ref{fig:qualitative_256} illustrates the visual differences between baseline methods and their CSF-Net–integrated versions.
Our method consistently produces sharper and more structurally coherent inpainting results across various mask types.
While ASUKA generates perceptually plausible outputs, it often suffers from object hallucination under large-mask conditions due to the lack of explicit semantic guidance.
By leveraging semantic cues and surrounding structural context, CSF-Net alleviates this issue through its fusion strategy, resulting in more realistic and semantically faithful completions.

\subsection{Ablation Study}
\label{sec:Ablation_Study}
All ablation studies are conducted using the LaMa under the Center Box (80\%) masking condition to ensure consistent analysis.
In Sec.~\ref{sec:4.1.Methods_Candidate_Generation}, we proposed a candidate generation based on Amodal Completion and selected the most reliable candidates using consistency scores.
Table~\ref{tab:ablation_candidate_score} shows the inpainting performance under different scoring strategies, showing that the combined metric $S_{\text{MSE}} + S_{\text{LPIPS}}$ achieves the best overall performance.

We investigate the effect of varying the number of candidate completions ($P$) in CSF-Net.
As shown in Tab.~\ref{tab:ablation_topk}, using $P$=3 achieves the best performance across all metrics.
Increasing $P$ to 5 or 10 results in degraded inpainting performance.
This is because top-ranked candidates already provide sufficient semantic diversity, while adding more may introduce redundant or low-quality results that interfere with the fusion process.
We also compare the case of $P$=1, which directly selects the top-ranked amodal completion without going through the Fusion Transformer and pixel selection process in CSF-Net.
Simply using the top-ranked candidate without fusion leads to lower performance in all metrics.
This shows that the fusion process in CSF-Net is essential for generating better and coherent results.

We further examine the effect of the Pixel Selection described in Sec.~\ref{sec:4.3.Methods_Pixel_Selection}.
As shown in Tab.~\ref{tab:ablation_Pixel_Selection}, incorporating pixel selection improves all evaluation metrics, demonstrating its importance in refining the guidance image with spatially coherent content.
Finally, we evaluate the impact of the encoder design proposed in Sec.~\ref{sec:4.2.Methods_CSF-Transformer}.
We tested a single-encoder design that takes the concatenated inputs $[\mathbf{I}_{\text{mask}}, \mathcal{H}_{\text{sel}}]$ along the channel dimension.
As shown in Tab.~\ref{tab:ablation_encoder}, the dual-encoder design (i.e., separate context and semantic branches) significantly outperforms the single-encoder across all metrics.
This confirms that separating contextual and candidate inputs enables better semantic disentanglement and leads to improved feature representations.

\begin{table}[t]
\resizebox{\columnwidth}{!}{%
\centering
\begin{tabular}{r|ccc}
\hline
Scores for Candidate Generation           & FID$\downarrow$ & LPIPS$\downarrow$ & C@m$\uparrow$ \\
\hline
MSE only $(S_{\text{MSE}})$                     & 25.91          & 0.379          & 0.676           \\
LPIPS only $(S_{\text{LPIPS}})$                 & 18.82          & 0.361          & 0.688           \\
MSE+LPIPS $(S_{\text{MSE}}+S_{\text{LPIPS}})$   & \textbf{15.93} & \textbf{0.326} & \textbf{0.701}  \\
\hline
\end{tabular}
}
\caption{Effect of consistency scores for candidate selection.}
\label{tab:ablation_candidate_score}
\end{table}

\begin{table}[t]
\centering
\small
\renewcommand{\arraystretch}{0.85}
\begin{tabular}{c|ccc}
\hline
\# of candidates ($P$) & FID$\downarrow$ & LPIPS$\downarrow$ & C@m$\uparrow$  \\ \hline
$P$ = 1  & 18.67          & 0.411          & 0.679          \\
$P$ = 3  & \textbf{15.93} & \textbf{0.326} & \textbf{0.701} \\
$P$ = 5  & 16.37          & 0.394          & 0.690          \\
$P$ = 10 & 16.51          & 0.406          & 0.682          \\
\hline
\end{tabular}
\caption{Ablation study on the number of candidate completions ($P$) and the impact of fusion in CSF-Net.}
\label{tab:ablation_topk}
\end{table}

\section{Conclusions and Future Work}
\label{sec:conclusions_and_limitations}
In this paper, We proposed CSF-Net, a transformer-based fusion framework that introduces object-level semantic guidance into the image inpainting process. 
By leveraging a pretrained amodal completion model, CSF-Net generates multiple structure-aware semantic candidates, which are fused with contextual information to produce a semantic guidance image ($\mathbf{I}_{\text{guide}}$).
This guidance enables more accurate and semantically aligned inpainting, particularly in challenging large-mask scenarios where contextual cues are limited.
CSF-Net can be seamlessly integrated into various inpainting backbones without any architectural modifications, demonstrating both its generality and practicality for real-world applications.
Extensive experiments on the \texttt{Places365}~\cite{zhou2017places} and \texttt{COCOA}~\cite{zhu2017semantic} datasets demonstrate that CSF-Net consistently improves performance across multiple inpainting baselines and evaluation metrics.

While CSF-Net shows strong results and wide compatibility, there are still areas for improvement.
First, the performance of CSF-Net can be affected by the quality of the amodal completion candidates. 
Although we apply a filtering step (Sec.~\ref{sec:4.1.Methods_Candidate_Generation}) to remove noisy or irrelevant outputs, the model may struggle when all candidates lack meaningful semantic information.
Second, it is used independently from the amodal completion and inpainting models. 
This modular design makes it easy to apply to other systems, but jointly training all components could improve the  results.

\begin{table}[t!]
\centering
\small
\renewcommand{\arraystretch}{0.85}
\begin{tabular}{l|ccc}
\hline
Encoder design & FID$\downarrow$ & LPIPS$\downarrow$ & C@m$\uparrow$ \\
\hline
w/o Pixel Selection & 18.54          & 0.421          & 0.684 \\
Ours   & \textbf{15.93} & \textbf{0.326} & \textbf{0.701} \\
\hline
\end{tabular}
\caption{Ablation study on Pixel Selection in CSF-Net.}
\label{tab:ablation_Pixel_Selection}
\end{table}

\begin{table}[t!]
\centering
\small
\renewcommand{\arraystretch}{0.85}
\begin{tabular}{l|ccc}
\hline
Encoder design & FID$\downarrow$ & LPIPS$\downarrow$ & C@m$\uparrow$ \\
\hline
Single-encoder & 21.08          & 0.430          & 0.689 \\
Ours   & \textbf{15.93} & \textbf{0.326} & \textbf{0.701} \\
\hline
\end{tabular}
\caption{Ablation study on encoder design in CSF-Net.}
\label{tab:ablation_encoder}
\end{table}

{
    \small
    \bibliographystyle{ieeenat_fullname}
    \bibliography{main}
}

\end{document}